%% file: 00-Main.tex
\documentclass{article}
\usepackage{amsmath,graphicx,mlspconf}
\input{math_commands.tex}

\usepackage{microtype}
\usepackage{subfigure}
\usepackage{booktabs} 
\usepackage{url}
\usepackage{cuted}
\usepackage{capt-of}
\usepackage{bm}
\usepackage{amssymb}
\usepackage{algorithm}
\usepackage{algorithmic}

\copyrightnotice{Preprint. Final published version can be found at: \url{https://doi.org/10.1109/MLSP49062.2020.9231618}}

\toappear{2020 IEEE International Workshop on Machine Learning for Signal Processing, Sept.\ 21--24, 2020, Espoo, Finland}

\title{PPO-CMA: Proximal Policy Optimization with Covariance Matrix Adaptation}
\name{%
    Perttu H\"am\"al\"ainen$^{\star}$%
    \qquad Amin Babadi$^{\star}$%
    \qquad Xiaoxiao Ma$^{\star}$%
    \qquad Jaakko Lehtinen$^{\star \dagger}$%
}
\address{%
    $^{\star}$ Aalto University, Helsinki, Finland \\%
    $^{\dagger}$ NVIDIA
}

\begin{document}
\ninept

\maketitle

\input{01-Abstract}

\input{02-Introduction}

\input{07-RelatedWork}

\input{03-Preliminaries}

\input{04-VarianceAdaptation}

\input{05-PPO-CMA}

\input{06-Evaluation}
\input{08-Conclusion} 

\bibliographystyle{IEEEbib}
\bibliography{10-References}

\input{09-Appendix}

\end{document}

%% file: math_commands.tex
\usepackage{amsmath,amsfonts,bm}

\def\eqref#1{equation~\ref{#1}}

\def\1{\bm{1}}

\DeclareMathAlphabet{\mathsfit}{\encodingdefault}{\sfdefault}{m}{sl}
\SetMathAlphabet{\mathsfit}{bold}{\encodingdefault}{\sfdefault}{bx}{n}

\newcommand{\sigmoid}{\sigma}



%% file: 01-Abstract.tex

\begin{abstract}
Proximal Policy Optimization (PPO) is a highly popular model-free reinforcement learning (RL) approach. However, we observe that in a continuous action space, PPO can prematurely shrink the exploration variance, which leads to slow progress and may make the algorithm prone to getting stuck in local optima. Drawing inspiration from CMA-ES, a black-box evolutionary optimization method designed for robustness in similar situations, we propose PPO-CMA, a proximal policy optimization approach that adaptively expands the exploration variance to speed up progress.  With only minor changes to PPO, our algorithm considerably improves performance in Roboschool continuous control benchmarks. Our results also show that PPO-CMA, as opposed to PPO, is significantly less sensitive to the choice of hyperparameters, allowing one to use it in complex movement optimization tasks without requiring tedious tuning.
\end{abstract}


%% file: 02-Introduction.tex

\section{Introduction}
Policy optimization with continuous state and action spaces is a central, long-standing problem in robotics and computer animation. In the general case, one does not have a differentiable model of the dynamics and must proceed by trial and error, i.e., try something (sample actions from  an exploration distribution, e.g., a neural network policy conditioned on the current state), see what happens, and learn from the results (update the exploration distribution such that good actions become more probable). In recent years, such approaches have achieved remarkable success in previously intractable tasks such as real-time locomotion control of (simplified) biomechanical models of the human body \cite{bergamin2019drecon}. One of the most popular policy optimization algorithms to achieve this is Proximal Policy Optimization (PPO) \cite{bergamin2019drecon,peng2018deepmimic,schulman2017proximal}.

\begin{figure}[t!]
\begin{center}
\includegraphics[width=3.3in]{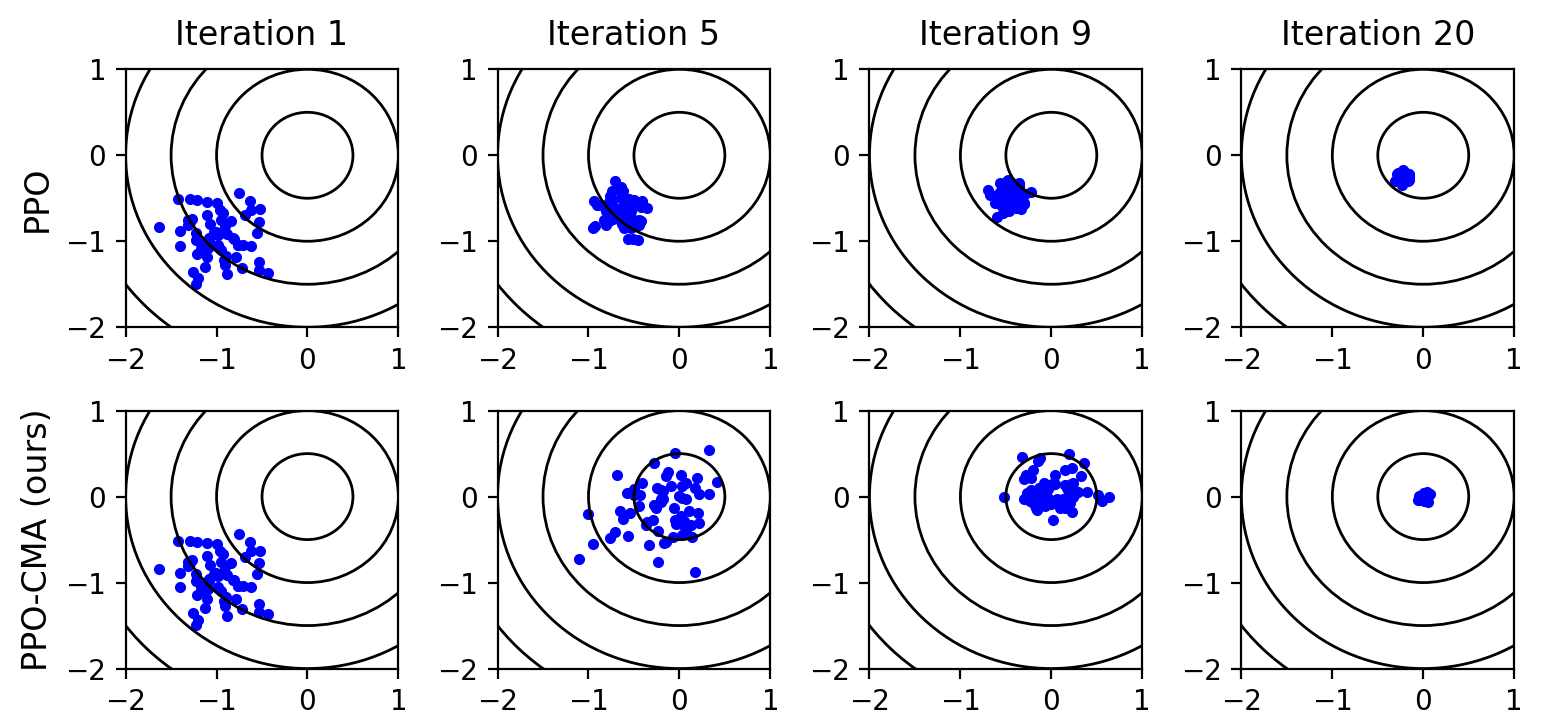}
\caption{Comparing PPO and PPO-CMA with a simple "stateless" quadratic objective. Sampled actions $\mathbf{a} \in \mathbb{R}^2$ are shown in blue. PPO shrinks the sampling/exploration variance prematurely, which leads to slow final progress. The proposed PPO-CMA method dynamically expands the variance to speed up progress, and only shrinks the variance when close to the optimum. Source code and animated visualization can be found at: \protect\url{https://github.com/ppocma/ppocma}.} \label{fig:teaser}
\end{center}
\end{figure}

In this paper, we make the following contributions:

\begin{itemize}
\item We provide novel evidence of how PPO's exploration variance can shrink prematurely, which leads to slow progress. Figure \ref{fig:teaser} illustrates this in a simple didactic problem. 

\item We propose PPO-CMA, a method that dynamically expands and contracts the exploration variance, inspired by the Covariance Matrix Adaptation Evolution Strategy (CMA-ES) optimization method. This only requires minor changes to vanilla PPO but improves performance considerably. 
\end{itemize}

%% file: 07-RelatedWork.tex

\section{Related Work}\label{sec:related}
Our work is closely related to Continuous Actor Critic Learning Automaton (CACLA) \cite{van2007reinforcement}. CACLA uses the sign of the advantage estimate -- in their case the TD-residual -- in the updates, shifting policy mean towards actions with positive sign. The paper also observes that using actions with negative advantages can have an adverse effect. However, using only positive advantage actions guarantees that the policy stays in the proximity of the collected experience. Thus, CACLA can be viewed as an early PPO approach, which we extend with CMA-ES -style variance adaptation. 


PPO represents on-policy RL methods, i.e., experience is assumed to be collected on-policy and thus must be discarded after the policy is updated. Theoretically, off-policy RL should allow better sample efficiency through the reuse of old experience, often implemented using an experience replay buffer. PPO-CMA can be considered as a hybrid method, since the policy mean is updated using on-policy experience, but the history or replay buffer for the variance update also includes older off-policy experience.


CMA-ES has been also applied to continuous control in the form of trajectory optimization. In this case, one can use CMA-ES to search for a sequence of optimal controls given an initial state \cite{babadi2018intelligent}. Although trajectory optimization approaches have demonstrated impressive results with complex humanoid characters, they require more computing resources in run-time. Trajectory optimization has also been leveraged to inform policy search using the principle of maximum entropy control \cite{levine2013guided}, which leads to a Gaussian policy. Furthermore, Differential Dynamic Programming (DDP) has been formulated in the terms of Gaussian distributions, which permits using CMA-ES for sampling the actions of each timestep in trajectory optimization \cite{rajamaki2018regularizing}. 

PPO-CMA is perhaps most closely related to \cite{abdolmaleki2018maximum,abdolmaleki2018relative}. Maximum a posteriori Policy Optimization (MPO) \cite{abdolmaleki2018maximum} also fits the policy to the collected experience using a weighted maximum likelihood approach, but negative weights are avoided through exponentiated Q-values, based on the control-inference dualism, instead of our negative advantage mirroring. Concurrently with our work, MPO has also been extended with decoupled optimization of policy mean and variance, yielding similar variance adaptation behaviour as CMA-ES and PPO-CMA; on a quadratic objective, variance is first increased and shrinks only when close to the optimum \cite{abdolmaleki2018relative}. 

Finally, it should be noted that PPO-CMA falls in the domain of model-free reinforcement learning approaches. In contrast, there are several model-based methods that learn approximate models of the simulation dynamics and use the models for policy optimization, potentially requiring less simulated or real experience. Both ES and RL approaches can be used for the optimization. Model-based algorithms are an active area of research, with recent work demonstrating excellent results in limited MuJoCo benchmarks \cite{chua2018deep}, but model-free approaches still dominate the most complex continuous problems such as humanoid movement.  


%% file: 03-Preliminaries.tex

\section{Preliminaries}
\subsection{Reinforcement Learning}\label{sec:RL}
We consider the discounted formulation of the policy optimization problem, following the notation of \cite{schulman2015high}. At time $t$, the agent observes a state vector $\mathbf{s}_t$ and takes an action $\mathbf{a}_t \sim \pi_\theta(\mathbf{a}_t | \mathbf{s}_t)$, where $\pi_\theta$ denotes the policy parameterized by $\mathbf{\theta}$, e.g., neural network weights. We focus on on-policy methods where the optimized policy also defines the exploration distribution. Executing the sampled action results in observing a new state $\mathbf{s}_t'$ and receiving a scalar reward $r_t$. The goal is to find $\mathbf{\theta}$ that maximizes the expected future-discounted sum of rewards $\mathbb{E}[\sum_{t=0}^{\infty}\gamma^t r_t]$, where $\gamma$ is a discount factor in the range $[0,1]$. A lower $\gamma$ makes the learning prefer instant gratification instead of long-term gains.

Both PPO and the PPO-CMA collect experience tuples $[\mathbf{s}_i,\mathbf{a}_i,r_i,\mathbf{s}_i']$ by simulating a number of \textit{episodes} in each optimization iteration. For each episode, an initial state $\mathbf{s_0}$ is sampled from some application-dependent stationary distribution, and the simulation is continued until a terminal (absorbing) state or a predefined maximum episode length $T$ is reached. After the iteration simulation budget $N$ is exhausted, $\mathbf{\theta}$ is updated. 


\subsection{Policy Gradient with Advantage Estimation}
Policy gradient methods update policy parameters by estimating the gradient $\mathbf{g}=\nabla_\theta \mathbb{E}[\sum_{t}^{\infty}\gamma^t r_t]$. PPO utilizes the following policy gradient loss:

\begin{equation}
\mathcal{L}_\theta=-\frac{1}{M}\sum_{i=1}^M A^\pi(\mathbf{s}_i,\mathbf{a}_i) \log \pi_\theta(\mathbf{a}_i | \mathbf{s}_i),\label{eq:pgLoss}
\end{equation}

where $i$ denotes minibatch sample index and $M$ is minibatch size. $A^\pi(\mathbf{s}_i,\mathbf{a}_i)$ denotes the \emph{advantage function}, which measures the benefit of taking action $\mathbf{a}_i$ in state $\mathbf{s}_i$. Positive $A^\pi$ means that the action was better than average and minimizing the loss function will increase the probability of sampling the same action again. Note that $A^\pi$ does not directly depend on $\theta$ and thus acts as a constant when computing the gradient of Equation \ref{eq:pgLoss}. A maximum likelihood view of Equation \ref{eq:pgLoss} is that the policy distribution is fitted to the data, each data point weighted by its advantage \cite{AWRPeng19}. Same as PPO, we use Generalized Advantage Estimation (GAE) \cite{schulman2015high}, a simple but effective way to estimate $A^\pi$.

\subsection{Continuous Action Spaces}
With a continuous action space, it is common to use a Gaussian policy. In other words, the policy network outputs state-dependent mean $\mathbf{\mu}_\theta(\mathbf{s})$ and covariance $\mathbf{C}_\theta(\mathbf{s})$ for sampling the actions. The covariance defines the exploration-exploitation balance. In the most simple case of isotropic unit Gaussian exploration, $\mathbf{C}= \mathbf{I}$, the loss function in Equation \ref{eq:pgLoss} becomes:

\begin{equation}
\mathcal{L}_\theta=\frac{1}{M}\sum_{i=1}^M A^\pi(\mathbf{s}_i,\mathbf{a}_i) || \mathbf{a}_i -  \mathbf{\mu}_\theta(\mathbf{s}_i)||^2,\label{eq:isotropicGaussianLoss}
\end{equation}

Intuitively, minimizing the loss drives the policy mean towards positive-advantage actions and away from negative-advantage actions. 

Following the original PPO paper, we use a diagonal covariance matrix parameterized by a vector $\mathbf{c}_\theta(\mathbf{s})=diag (\mathbf{C}_\theta(\mathbf{s}))$. In this case, the loss becomes:


\begin{eqnarray}
\mathcal{L}_\theta&=&\frac{1}{M}\sum_{i=1}^M A^\pi(\mathbf{s}_i,\mathbf{a}_i)\sum_{j} \Big[\frac{(a_{i,j} -  \mu_{j;\theta}(\mathbf{s}_i) )^2}{c_{j;\theta}(\mathbf{s}_i)} \nonumber \\ 
&& + 0.5 \log c_{j;\theta}(\mathbf{s}_i)\Big],\label{eq:gaussianLoss}
\end{eqnarray}
where $i$ indexes over a minibatch and $j$ indexes over action variables.  

\subsection{Proximal Policy Optimization}
The basic idea of PPO is that one performs not just one but multiple minibatch gradient steps with the experience of each iteration. Essentially, one reuses the same data to make more progress per iteration, while stability is ensured by limiting the divergence between the old and updated policies \cite{schulman2017proximal}. PPO is a simplification of Trust Region Policy Optimization (TRPO) \cite{schulman2015trust}, which uses a more computationally expensive approach to achieve the same.

The original PPO paper \cite{schulman2017proximal} proposes two variants: 1) using an additional loss term that penalizes KL-divergence between the old and updated policies, and 2) using the so-called clipped surrogate loss function. The paper concludes that the clipped surrogate loss is the recommended choice. This is also the version that we use in this paper in all PPO vs. PPO-CMA comparisons. 

%% file: 04-VarianceAdaptation.tex
\section{Problem Analysis and Visualization}
To allow simple visualization of actions sampled from the policy, the didactic problem in Figure \ref{fig:teaser} simplifies policy optimization into a generic black-box optimization problem in a 2D action space. 

To achieve this, we set $\gamma=0$, which simplifies the policy optimization objective $\mathbb{E}[\sum_{t=0}^{\infty}\gamma^t r_t]=\mathbb{E}[r_0]=\mathbb{E}[r(\mathbf{s},\mathbf{a})]$, where $\mathbf{s}$ and $\mathbf{a}$ denote the first state and action of an episode. Thus, we can use $T=1$ and focus on visualizing only the first timesteps of each episode. Further, we use a state-agnostic $r(\mathbf{s},\mathbf{a})=r(\mathbf{a})=-\mathbf{a}^T\mathbf{a}$. Thus, we have a simple quadratic optimization problem and it is enough to only visualize the action space. The optimal policy Gaussian has zero mean and variance. 

\subsection{The Instability Caused by Negative Advantages}
\begin{figure}[t]
\begin{center}
\includegraphics[width=3.2in]{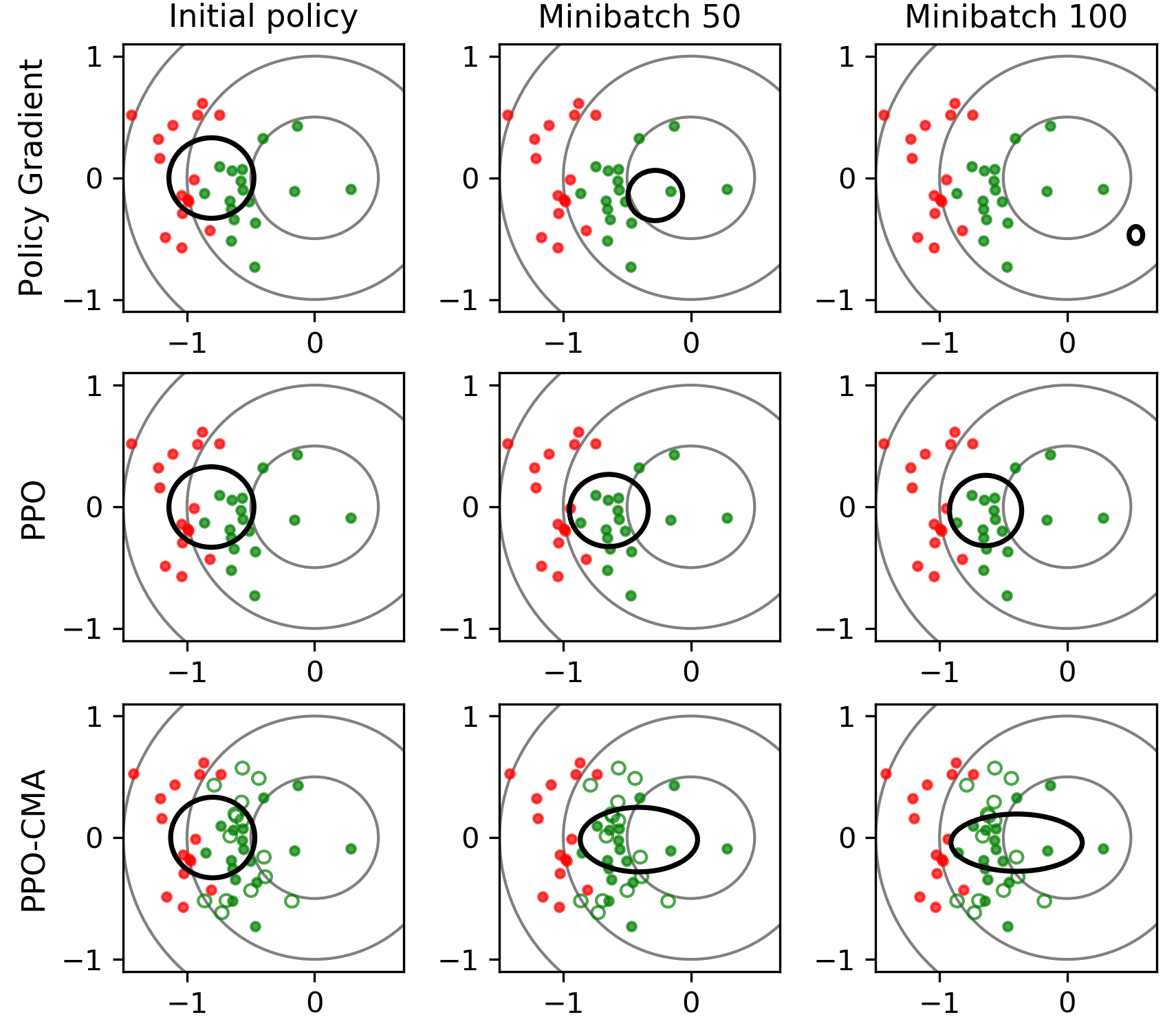}
\end{center}
\caption{Policy evolution over the minibatch gradient steps of a single iteration in our didactic problem. Figure \ref{fig:teaser} shows how the same methods perform over multiple iterations. The black ellipses denote the policy mean and standard deviation according to which actions are sampled/explored. Positive-advantage actions are shown in green, negative advantages in red. The green non-filled circles show the negative-advantage actions converted to positive ones (Section \ref{sec:mirroring}). Policy gradient diverges. PPO limits the update before convergence or divergence. PPO-CMA expands the variance in the progress direction, improving exploration in subsequent iterations.}\label{fig:pg}
\end{figure}

Considering the Gaussian policy gradient loss functions in Equations \ref{eq:isotropicGaussianLoss} and \ref{eq:gaussianLoss}, we note a fundamental problem: \emph{Actions with a negative advantages may cause instability} when performing multiple minibatch gradient steps in PPO style, as each step drives the policy Gaussian further away from the negative-advantage actions. This is visualized in Figure \ref{fig:pg} (top row). The policy diverges, gravitating away from the negative-advantage actions. 

Note that policy gradient does not diverge if one collects new experience and re-estimates the advantages between each gradient step; this, however, is what PPO aims to avoid for better sample efficiency.

\subsection{PPO: Stable But May Converge Prematurely}
%

As visualized in Figure \ref{fig:teaser}, the clipped surrogate loss of PPO prevents divergence, but the exploration variance can shrink prematurely when performing the updates over multiple iterations. Possibly to mitigate this, PPO also adds an entropy loss term that penalizes low variance. However, the weight of the entropy loss can be difficult to finetune. 

Although the original PPO paper \cite{schulman2017proximal} demonstrated good results in MuJoCo problems with a Gaussian policy, the most impressive Roboschool results did not adapt the variance through gradient updates. Instead, the policy network only output the Gaussian mean and a linearly decaying variance with manually tuned decay rate was used. Thus, our observation of variance adaptation issues complements their work instead of contradicting it.

\subsection{How CMA-ES Solves the Problems}\label{sec:cmaes}
The Evolution Strategies (ES) literature has a long history of solving similar Gaussian exploration and variance adaptaion problems, culminating in the widely used CMA-ES optimization method and its recent variants  \cite{hansen2006cma,loshchilov2017limited}. CMA-ES is a black-box optimization method for finding a parameter vector $\mathbf{x}$ that maximizes some objective or fitness function $f(\mathbf{x})$. The key difference to PPO is that the exploration distribution's mean and covariance are plain variables instead of neural  networks, and not conditioned on an agent's current state. However, as we propose in Section \ref{sec:PPOCMA}, one can adapt the key ideas of CMA-ES to policy optimization.  

The CMA-ES core iteration is summarized in Algorithm \ref{alg:CMAES}. Although there is no convergence guarantee, CMA-ES performs remarkably well on multimodal and/or noisy functions if using enough samples per iteration. For full details of the update rules, the reader is referred to Hansen's excellent tutorial \cite{hansen2016cma}.

\begin{algorithm}[h]
\begin{algorithmic}[1]
\FOR{iteration=1,2,...}
\STATE Draw samples $\mathbf{x}_i \sim \mathcal{N}(\bm{\mu},\mathbf{C})$.
\STATE Evaluate $f(\mathbf{x}_i)$.
\STATE Sort the samples based on $f(\mathbf{x}_i)$ and compute weights $\mathbf{w}_i$ based on the ranks, such that best samples have highest weights.
\STATE Update $\bm{\mu}$ and $\mathbf{C}$ using the samples and weights.
\ENDFOR
\end{algorithmic}
\caption{High-level summary of CMA-ES}\label{alg:CMAES}
\end{algorithm}

CMA-ES avoids premature convergence and instability through the following:
\begin{enumerate}
\item When fitting the sampling distribution to weighted samples, only positive weights are used.  
\item The so-called rank-$\mu$ update and evolution path heuristic elongate the exploration variance in the progress direction, making premature convergence less likely.
\end{enumerate} 

Below, we overview these key ideas, providing a foundation for Section \ref{sec:PPOCMA}, which adapts the ideas for episodic RL. Note that CMA-ES is not directly applicable to RL, as instead of a single action optimization task, RL is in effect solving multiple action optimization tasks in parallel, one for each possible state. With a continuous state space, one cannot enumerate the samples for each state, which means that the sorting operation of Algorithm \ref{alg:CMAES} is not feasible.

\subsubsection{Computing Sample Weights}\label{sec:pruning_and_weights}
Using the default CMA-ES parameters, the weights of the worst 50\% of samples are set to 0, i.e., samples below median fitness are pruned and have no effect. The exploration mean $\bm{\mu}$ is updated in maximum likelihood manner to weighted average of the samples (similar to minimizing the loss in Equation \ref{eq:isotropicGaussianLoss} with the advantages as the weights). Because CMA-ES uses only non-negative weights, the maximum likelihood update does not diverge from the sampled actions. This suggests that one could similarly only use positive advantage actions (the better half of the explored actions) in PPO.


\subsubsection{The rank-$\mu$ update}\label{sec:cmarankmu}
\begin{figure}[h]
\begin{center}
\includegraphics[width=0.9\linewidth]{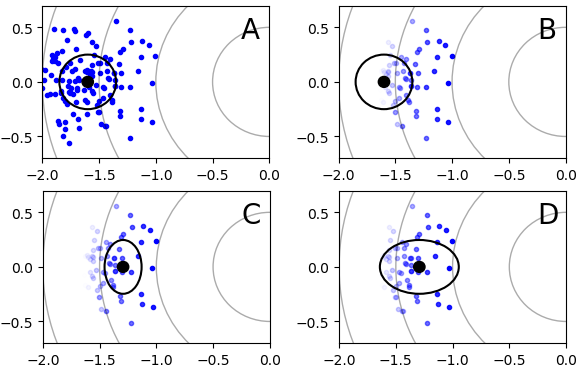}
\end{center}
\caption{The difference between joint and separate updating of mean and covariance, denoted by the black dot and ellipse. A) sampling, B) pruning and weighting of samples based on fitness, C) EMNA-style update, i.e., estimating mean and covariance based on weighted samples, D) CMA-ES rank-$\mu$ update, where covariance is estimated before updating the mean. This elongates the variance in the progress direction, improving exploration in the next iteration. }\label{fig:cmaVsEmna}
\end{figure}

Superficially, the core iteration loop of CMA-ES is similar to other optimization approaches with recursive sampling and distribution fitting such as the Cross-Entropy Method \cite{de2005tutorial} and Estimation of Multivariate Normal Algorithm (EMNA) \cite{larranaga2001estimation}. However, there is a crucial difference: in the so-called Rank-$\mu$ update, \textit{CMA-ES first updates the covariance and only then updates the mean} \cite{hansen2016cma}. This has the effect of elongating the exploration distribution along the best search directions instead of shrinking the variance prematurely, as shown in Figure \ref{fig:cmaVsEmna}. 

\subsubsection{Evolution path heuristic}\label{sec:evpathcma}
CMA-ES also features the so-called evolution path heuristic, where a component $\alpha \mathbf{p}^{(i)} \mathbf{p}^{(i)T}$ is added to the covariance, where $\alpha$ is a scalar, the $(i)$ superscript denotes iteration index, and $\mathbf{p}$ is the evolution path \cite{hansen2016cma}:

\begin{equation}
\mathbf{p}^{(i)}=\beta_0 \mathbf{p}^{(i-1)}+\beta_1 (\bm{\mu}^{(i)} - \bm{\mu}^{(i-1)}). \label{eq:evpath}
\end{equation}

Although the exact computation of the default $\beta_0$ and $\beta_1$ multipliers is rather involved, Equation \ref{eq:evpath} essentially amounts to first-order low-pass filtering of the steps taken by the distribution mean $\bm{\mu}$ between iterations. When CMA-ES progresses along a continuous slope of the fitness landscape, $||\mathbf{p}||$ is large, and the covariance is elongated and exploration is increased along the progress direction. Near convergence, when CMA-ES zigzags around the optimum in a random walk, $||\mathbf{p}|| \approx 0$ and the evolution path heuristic has no effect.

%% file: 05-PPO-CMA.tex
\section{PPO-CMA}\label{sec:PPOCMA}
Building on the previous section, we can now describe our PPO-CMA method, summarized in Algorithm \ref{alg:PPOCMA}. PPO-CMA is simple to implement, only requiring the following changes to PPO:

\begin{itemize}
\item Instead of the clipped surrogate loss, we use the standard policy gradient loss in Equation \ref{eq:gaussianLoss} and train only on actions with positive advantage estimates to ensure stability, as motivated in Section \ref{sec:pruning_and_weights}. However, as setting negative advantages to zero discards information, we also propose a mirroring technique for converting negative-advantage actions to positive ones (Section \ref{sec:mirroring}). 
\item We implement the rank-$\mu$ update (Section \ref{sec:cmarankmu}) using separate neural networks for policy mean and variance. This way, the variance can be updated before updating the mean. 
\item We maintain a history of training data over $H$ iterations, used for training the variance network. This approximates the CMA-ES evolution path heuristic, as explained below. 
\end{itemize}
Together, these features result in the emergence of PPO-CMA variance adaptation behavior shown in Figures \ref{fig:teaser} and \ref{fig:pg}. Despite the differences to original PPO, we still consider PPO-CMA a proximal policy optimization method, as fitting the new policy to the positive-advantage actions prevents it diverging far from the current policy.  

\begin{algorithm}[b]
\caption{PPO-CMA}
\label{alg:PPOCMA}
\begin{algorithmic}[1]
\FOR{iteration=1,2,...}
\WHILE{iteration simulation budget $N$ not exceeded}
\STATE Reset the simulation to a (random) initial state
\STATE Run agent on policy $\pi_\theta$ for $T$ timesteps or until a terminal state 
\ENDWHILE
\STATE Train critic network for $K$ minibatches using the experience from the current iteration
\STATE Estimate advantages $A^\pi$ using GAE \cite{schulman2015high}
\STATE Clip negative advantages to zero, $A^\pi \leftarrow \max(A^\pi,0)$ \label{row:clip} or convert them to positive ones (Section \ref{sec:mirroring})
\STATE Train policy variance for $K$ minibatches using experience from past $H$ iterations and Eq. \ref{eq:gaussianLoss}
\STATE Train policy mean for $K$ minibatches using the experience from this iteration and Eq. \ref{eq:gaussianLoss} \label{row:trainMean}
\ENDFOR
\end{algorithmic}
\end{algorithm}


\subsection{Approximating the Evolution Path Heuristic}\label{sec:evpath}
We approximate the CMA-ES evolution path heuristic (Section \ref{sec:evpathcma}) by keeping a history of $H$ iterations of data and sampling the variance training minibatches from the history instead of only the latest data. Similar to the original evolution path heuristic, this elongates the variance for a given state if the mean is moving in a consistent direction. We do not implement the CMA-ES evolution path heuristic directly, because this would need yet another neural network to maintain and approximate a state-dependent $\mathbf{p}(\mathbf{s})$. Similar to exploration mean and variance, $\mathbf{p}$ is a CMA-ES algorithm state variable; in policy optimization, such variables become functions of agent state and need to be encoded as neural network weights.





\subsection{Mirroring Negative-Advantage Actions}\label{sec:mirroring}
Disregarding negative-advantage actions may potentially discard valuable information. We observe that assuming linearity of advantage around the current policy mean $\bm{\mu}(\mathbf{s}_i)$, it is possible to mirror negative-advantage actions about the mean to convert them to positive-advantage actions. More precisely, we set $\mathbf{a}_i'=2\bm{\mu}(\mathbf{s}_i)-\mathbf{a}_i, A^\pi(\mathbf{a}_i')=-A^\pi(\mathbf{a_i})\psi(\mathbf{a}_i,\mathbf{s}_i)$, where $\psi(\mathbf{a}_i,\mathbf{s}_i)$ is a Gaussian kernel (we use the same shape as the policy) that assigns less weight to actions far from the mean. This is visualized at the bottom of Figure \ref{fig:pg}. The mirroring drives the policy Gaussian away from worse than average actions, but in a way consistent with the weighted maximum likelihood estimation perspective which requires non-negative weights for stability. If the linearity assumption holds, the mirroring effectively doubles the amount of data informing the updates.

In the CMA-ES literature, a related technique is to use a negative covariance matrix update procedure \cite{hansen2010benchmarking}, but the technique does not improve the estimation of the mean.

%% file: 06-Evaluation.tex

\section{Evaluation}\label{sec:evaluation} 



A key issue in the usability of an RL method is sensitivity to hyperparameters. As learning complex tasks can take hours or days, finetuning hyperparameters is tedious. Thus, we conducted hyperparameter searches to investigate following questions: 
\begin{itemize}
\item Can PPO-CMA produce better results than PPO without precise tuning of hyperparameters?
\item Can hyperparameters optimized for simple tasks generalize to complex tasks? 
\item Is PPO-CMA less prone to getting stuck in local optima? 
\end{itemize}

As elaborated below, our data indicates a positive answer to all the questions. Furthermore, we conducted an ablation study that verifies that all the proposed PPO-CMA algorithm features improve the results.

We used the 9 2D OpenAI Gym Roboschool continuous control environments \cite{OpenAI_roboschool} for the hyperparameter searches and the ablation study and tested generalization on the OpenAI Gym MuJoCo Humanoid \cite{brockman2016openai}. We performed 5 training runs with different random seeds for environment and setting (total 31500 runs). The same codebase was used for both PPO and PPO-CMA, only toggling PPO-CMA features on/off, to ensure no differences due to code-level optimizations \cite{engstrom2019implementation}.



\subsection{Sensitivity to Hyperparameters}
\begin{figure}[t]
\begin{center}
\includegraphics[width=1.0\linewidth]{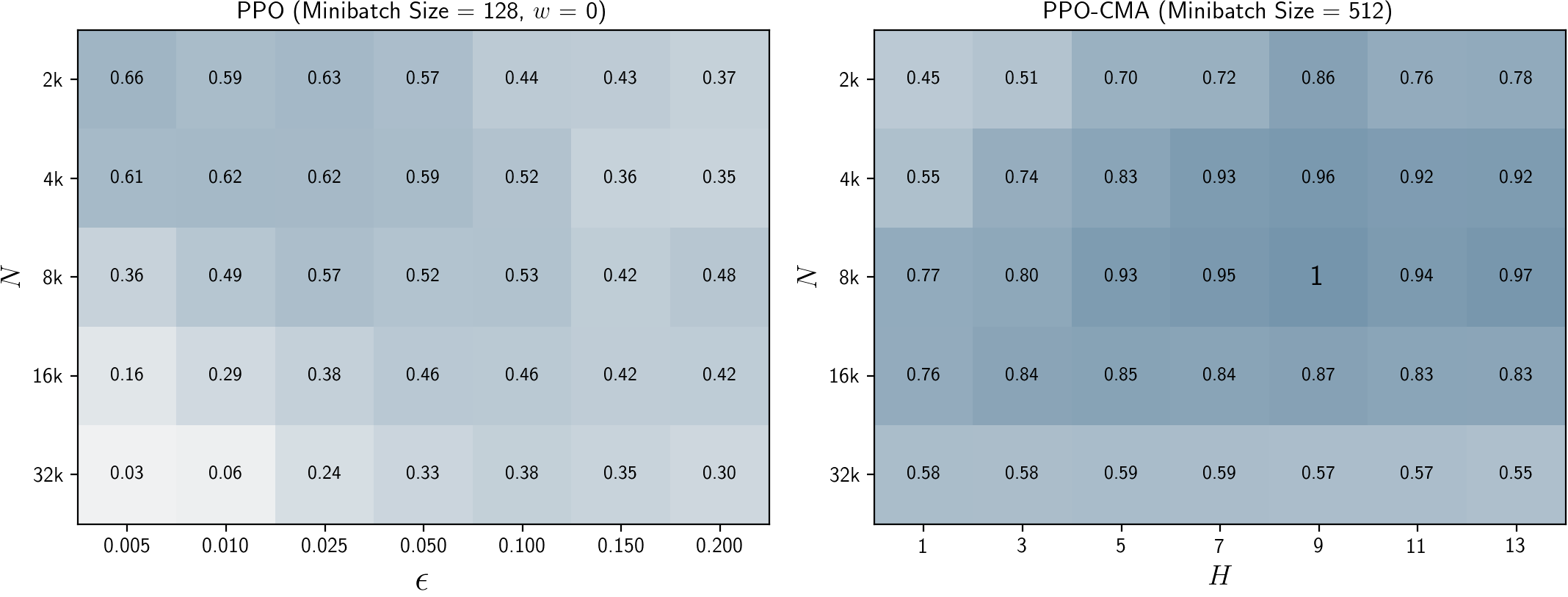}
\end{center}
\caption{PPO and PPO-CMA performance as a function of key hyperparameters, using average normalized scores from 9 Roboschool environments (1 is the best observed score). The batch sizes and entropy loss weight $w$ values are the ones that produced the best results in our hyperparameter searches. PPO-CMA performs overall better, is not as sensitive to the hyperparameter choices, and the hyperparameters can be adjusted more independently. In contrast, PPO requires careful finetuning of both the $\epsilon$ and $N$ parameters.}\label{fig:PPOvsPPO-CMA}
\end{figure}

Figure \ref{fig:PPOvsPPO-CMA} visualizes sensitivity to key hyperparameters, i.e., iteration simulation budget $N$, PPO-CMA's history buffer size $H$, and PPO's clipping parameter $\epsilon$ that determines how much the updated policy can diverge from the old one. The scores in Figure \ref{fig:PPOvsPPO-CMA} are normalized averages over all runs for a hyperparameter and algorithm choice, such that the best best parameter combination and algorithm has score 1. We tested minibatch sizes 128, 256, 512, 1024, 2048 and use the best performing ones for Figure \ref{fig:PPOvsPPO-CMA}. The figure reveals the following:

\begin{itemize}
\item PPO-CMA performs better with a wide range of hyperparameters, in particular with $H \ge 5$. Similar to CMA-ES, the main parameter to adjust is $N$. A large $N$ makes progress more robust and less noisy. On the other hand, a large $N$ means less iterations and possibly less progress within some total simulation budget, which shows as the lower scores for the largest $N$ in Figure \ref{fig:PPOvsPPO-CMA}.
\item There is a strong interaction of PPO's $\epsilon$ and $N$; if one is changed, the other must be also changed. This makes finetuning the parameters difficult, especially considering that PPO has the additional entropy loss weight parameter to tune. Interestingly, the optimal parameter combination appears to be a very low $\epsilon$ together with a very low $N$. On the other hand, $N$ should not be decreased below the episode time limit $T$. Most of the Roboschool environments use $T=1000$.
\end{itemize}

\subsection{Generalization and Scaling to Complex Problems}
\begin{figure}[h]
\begin{center}
\includegraphics[width=3.2in]{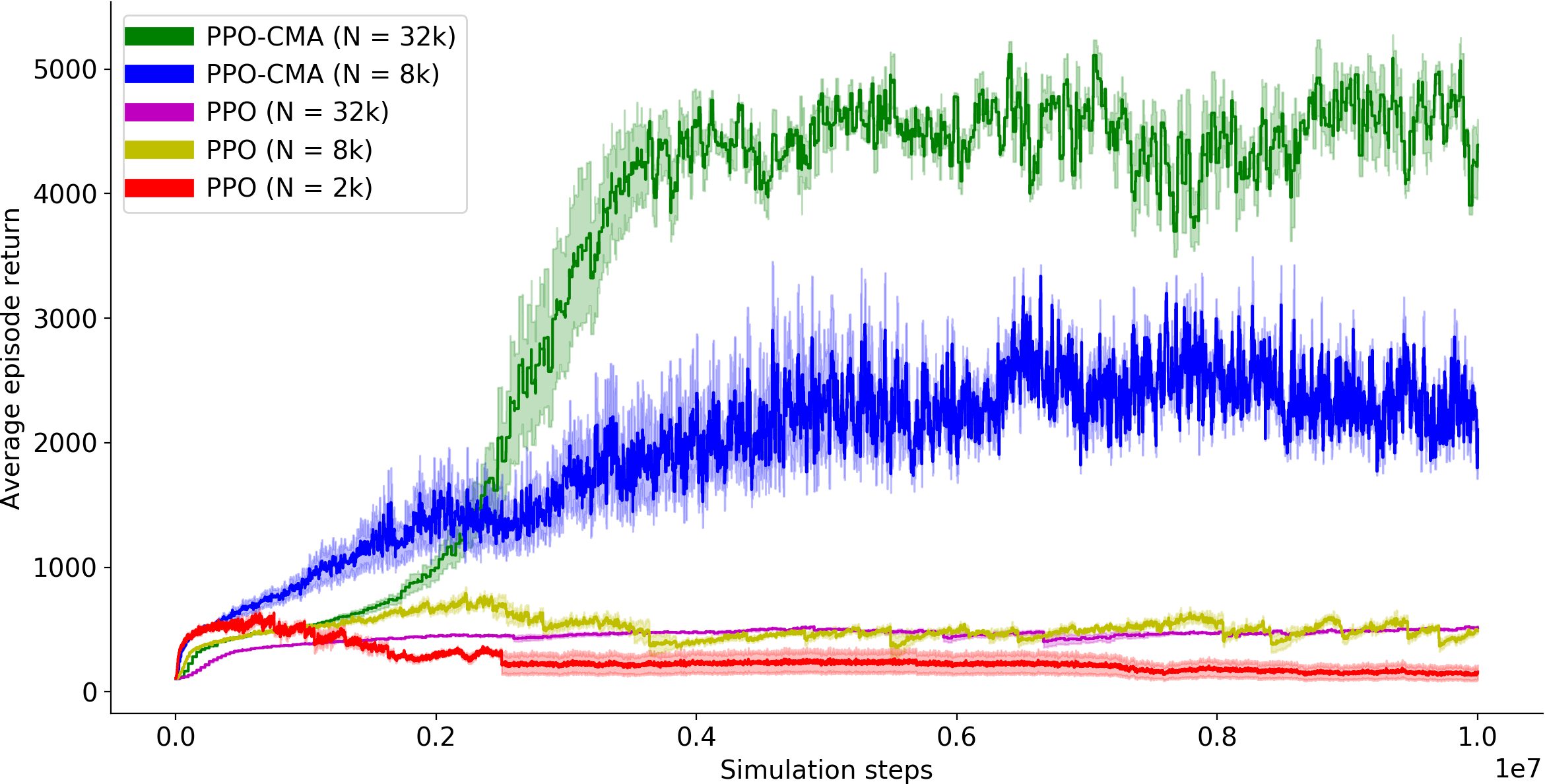}
\end{center}
\caption{Comparing PPO and PPO-CMA in the MuJoCo Humanoid-v2 environment, showing means and standard deviations of training curves from 3 runs with different random seeds.}\label{fig:HumanoidPlot}
\end{figure}

The 9 environments used for the hyperparameter search are all relatively simple 2D tasks. To test generalization and scaling, we ran both algorithms on the more challenging MuJoCo Humanoid-v2 environment \cite{brockman2016openai}. We used the best-performing hyperparameters of Figure \ref{fig:PPOvsPPO-CMA} and also tested a larger simulation budget $N$. The agent is a 3D humanoid that gets rewards for forward locomotion. The results are shown in Figure \ref{fig:HumanoidPlot}.  PPO-CMA yields clearly better results, especially with the increased $N$.

\subsection{Ablation study}
To ensure that PPO-CMA has no redundant features, we tested different ablated versions on the 9 Roboschool environments used for Figure \ref{fig:PPOvsPPO-CMA}. Table \ref{table:ablation} shows the resulting normalized scores. The results indicate that all the proposed algorithm components improve performance. 

\begin{table}[h]
\small
\begin{center}
\begin{tabular}{| c | c |}
\hline
\textit{Algorithm version} & \textit{Score} \\
\hline
No mirroring, no evolution path, no rank-$\mu$ & 0.57  \\
No mirroring, no evolution path heuristic & 0.71 \\
No mirroring of negative-advantage actions & 0.82 \\
Full PPO-CMA & 1 \\
\hline
\end{tabular}
\caption{Ablation study results, showing normalized scores similar to Figure \ref{fig:PPOvsPPO-CMA}.  Note that not using the rank-$\mu$ heuristic amounts to using a single policy network that outputs both mean and variance.}\label{table:ablation}
\end{center}
\end{table}

%% file: 08-Conclusion.tex
\section{Limitations}

We only compare PPO-CMA to PPO, and recognize that there are more recent RL algorithms that may offer better performance \cite{abdolmaleki2018maximum, abdolmaleki2018relative,chua2018deep}. However, PPO is a widely used algorithm due to its combination of simplicity and good performance, which calls for better understanding of its limitations. 


\section{Conclusion}
Proximal Policy Optimization (PPO) is a simple, powerful, and widely used model-free reinforcement learning approach. However, we have observed that PPO can prematurely shrink the exploration variance, leading to slow convergence. As a solution to the variance adaptation problem, we have proposed the PPO-CMA algorithm that adopts the rank-$\mu$ update and evolution path heuristics of CMA-ES to episodic RL.  

PPO-CMA improves PPO results in the tested Roboschool continuous control tasks while not sacrificing mathematical and conceptual simplicity. We add the separate neural networks for policy mean and variance and the $H$ hyperparameter, but on the other hand, we do not need PPO's clipped surrogate loss function, the $\epsilon$ parameter, or the entropy loss term. Similar to CMA-ES, PPO-CMA can be said to be quasi-parameter-free; according to our experience, once the neural network architecture is decided on, one mainly needs to increase the iteration sampling budget $N$ for more difficult problems.

\section*{Acknowledgments}
This work has been supported by Academy of Finland grants 299358 and 305737. 

%% file: 09-Appendix.tex

\appendix

\section{Implementation Details}
Similar to previous work, we use a fully connected policy network with a linear output layer and treat the variance output as log variance $\mathbf{v}=\log (\mathbf{c})$. In our initial tests with PPO, we ran into numerical precision errors which could be prevented by soft-clipping the mean as $\bm{\mu}_{clipped}= \mathbf{a}_{min} + (\mathbf{a}_{max} - \mathbf{a}_{min})\ \otimes \sigmoid(\bm{\mu})$, where $\mathbf{a}_{max}$ and $\mathbf{a}_{min}$ are the action space limits. Similarly, we clip the log variance as $\mathbf{v}_{clipped}=\mathbf{v}_{min} + (\mathbf{v}_{max} - \mathbf{v}_{min})\ \otimes \sigmoid(\mathbf{v)}$, where $\mathbf{v}_{min}$ is a lower limit parameter, and $\mathbf{v}_{max}=2\log(\mathbf{a}_{max}-\mathbf{a}_{min})$.  

We use a lower standard deviation limit of 0.01. Thus, the clipping only ensures numerical precision but has little effect on convergence. The clipping is not necessary for PPO-CMA in our experience, but we still use with both algorithms it to ensure a controlled and fair comparison. 

To ensure a good initialization, we pretrain the policy in supervised manner with randomly sampled observation vectors and a fixed target output $\mathbf{v}_{clipped}=2\log(0.5(\mathbf{a}_{max}-\mathbf{a}_{min}))$ and $\bm{\mu}_{clipped}=0.5(\mathbf{a}_{max}+\mathbf{a}_{min})$. The rationale behind this choice is that the initial exploration Gaussian should cover the whole action space but the variance should be lower than the upper clipping limit to prevent zero gradients. Without the pretraining, nothing quarantees sufficient exploration for all observed states.

We train both the policy and critic networks using Adam. Table \ref{table:hyperparams} lists all our hyperparameters not included in the hyperparameter searches. 

\begin{table}[H]
\small
\begin{center}
\begin{tabular}{| c | c |}
\hline
\textit{Hyperparameter} & \textit{Value} \\
\hline
Training minibatch steps per iteration ($K$) & 100 \\
Adam learning rate & 0.0003 \\
Network width & 128 \\
Number of hidden layers & 2  \\
Activation function & Leaky ReLU  \\
Action repeat & 2 \\
Critic loss & L1 \\
\hline
\end{tabular}
\caption{Hyperparameters used in our PPO and PPO-CMA implementations.}\label{table:hyperparams}
\end{center}
\end{table}

We use the same network architecture for all neural networks. Action repeat of 2 means that the policy network is only queried for every other simulation step and the same action is used for two steps. This speeds up training.  

We use L1 critic loss as it seems to make both PPO and PPO-CMA less sensitive to the reward scaling. For better tolerance to varying state observation scales, we use an automatic normalization scheme where observation variable $j$ is scaled by $k_{j}^{\left(i\right)}=min\left(k_{j}^{\left(i-1\right)},1/\left(\rho_{j}+\kappa\right) \right)$, where $\kappa=0.001$ and $\rho_{j}$ is the root mean square of the variable over all iterations so far. This way, large observations are scaled down but the scaling does not constantly keep adapting as training progresses.  

Following Schulman's original PPO code, we also use episode time as an extra feature for the critic network to help minimize the value function prediction variance arising from episode termination at the environment time limit. Note that as the feature augmentation is not done for the policy, this has no effect on the usability of the training results.

Our implementation trains the policy mean and variance networks in separate passes, keeping one network fixed while the other is trained. An alternative would be to train both networks at the same time, but cache the policy means and variances when sampling the actions, and use the cached mean for the variance network's loss function, and the cached variance for the mean network's loss.

\section{Hyperparameter Search Details}
\begin{figure*}[htb]
\begin{center}
\includegraphics[width=0.75\textwidth]{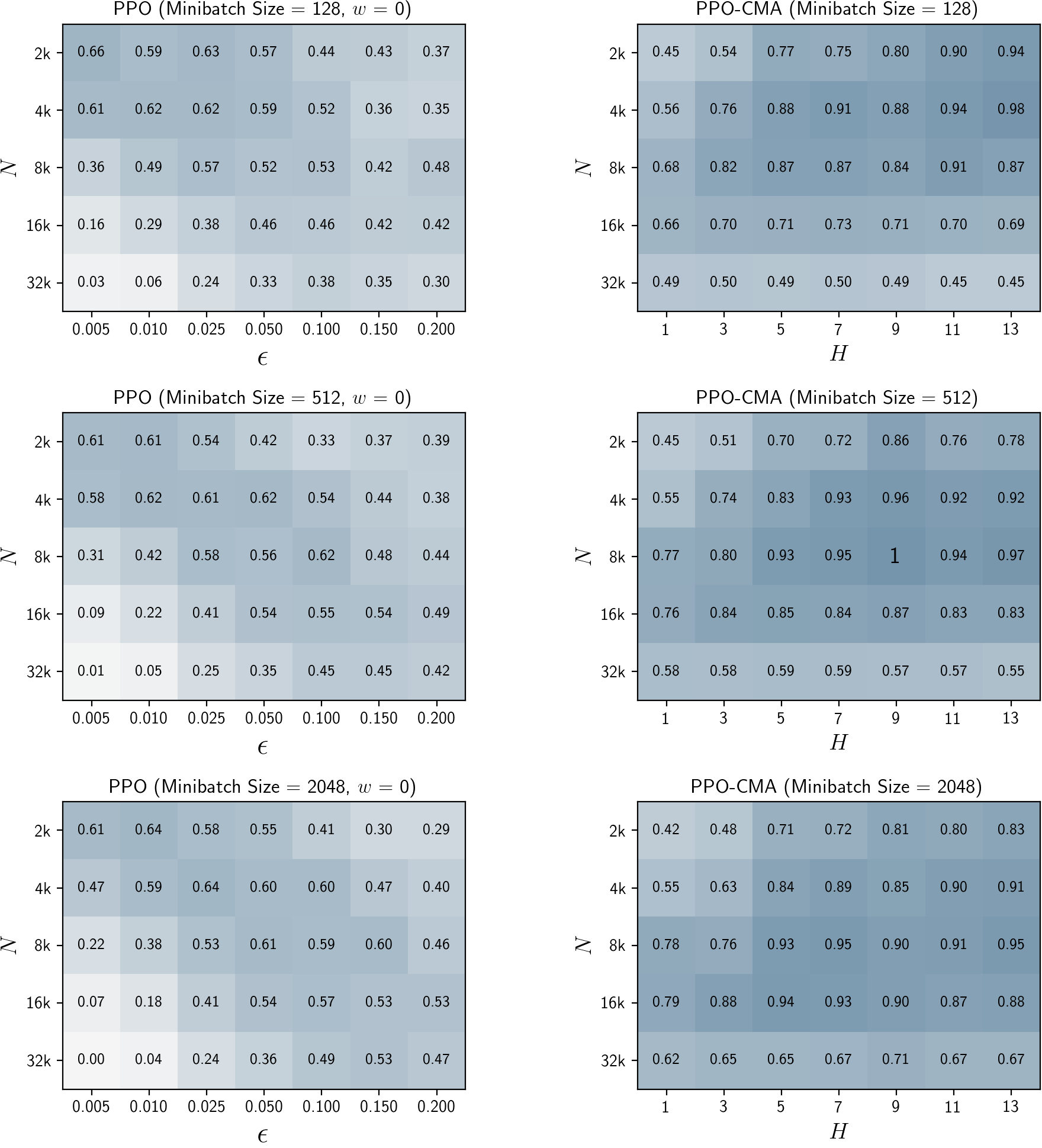}
\end{center}
\caption{Slices of 3D hyperparameter search spaces, comprising minibatch size, $N$, and $\epsilon$ for PPO, and minibatch size, $N$, and $H$ for PPO-CMA. Minibatch size has only minor effect, and the slices with different minibatch sizes look approximately similar.}\label{fig:GridPackPlot}
\end{figure*}

The PPO vs. PPO-CMA comparison of Section \ref{sec:evaluation} uses the best hyperparameter values that we found through an extensive search process. We performed the following searches:
\begin{enumerate}
\item A 3D search over PPO's $N$, $\epsilon$, and minibatch size, using the values in Figure \ref{fig:PPOvsPPO-CMA} plus minibatch sizes 128, 256, 512, 1024, 2048. 
\item A 3D search over PPO's $N$, $\epsilon$, and entropy loss weight $w \in \left\lbrace0, 0.01, 0.05, 0.1, 0.15\right\rbrace$, keeping the minibatch size at 128, which yielded the best results in the search above. Instead of a full 4D search, we chose this simplification to conserve computing resources and because the minibatch size was found to only have a minor effect on PPO's performance, as shown in Figure \ref{fig:GridPackPlot}.
\item A 3D search over PPO-CMA's $N$, $H$, and minibatch size, using the values in Figure \ref{fig:PPOvsPPO-CMA} plus minibatch sizes 128, 256, 512, 1024, 2048. 
\end{enumerate}

Each hyperparameter combination was tested using 5 independent training runs of the following 9 Roboschool environments: \textit{Inverted pendulum, Inverted pendulum swing-up, Inverted double pendulum, Reacher, Hopper, Walker2d, HalfCheetah, Ant,} and \textit{Pong.} In total, we performed 31500 training runs, totaling roughly two CPU years. 

Each hyperparameter combination score in Figure \ref{fig:PPOvsPPO-CMA} is the average of 45 normalized scores: 5 training runs with different random seeds up to 1M simulation steps, using the 9 OpenAI Gym Roboschool tasks. The scores were normalized as $R_{norm}=(R - R_{min})/(R_{max} - R_{min})$, where $R$ is a training run's average of the non-discounted episode return $\sum_{t} r_t$ from the last iteration, and $R_{min},R_{max}$ are the minimum and maximum $R$ of the same task over all training runs and tested hyperparameters. After the averaging, we perform a final normalization over all tested parameter combinations and algorithms such that the best combination and algorithm has score 1, i.e., PPO-CMA with minibatch size 512, $N=8k$, and $H=9$ in Figure \ref{fig:PPOvsPPO-CMA}. 

We did not have the computing resources to conduct the hyperparameter searches using all the Roboschool tasks such as the Atlas humanoid robot simulation; thus, we focused on the more simple 2D tasks that could be assumed to be solved within the limit of 1M simulation steps. The results in Figure \ref{fig:HumanoidPlot} indicate that the found parameters generalize beyond the simple tasks.


\section{Additional Results}\label{sec:additionalResults}
\begin{figure*}[tb]
\begin{center}
\includegraphics[width=0.75\textwidth]{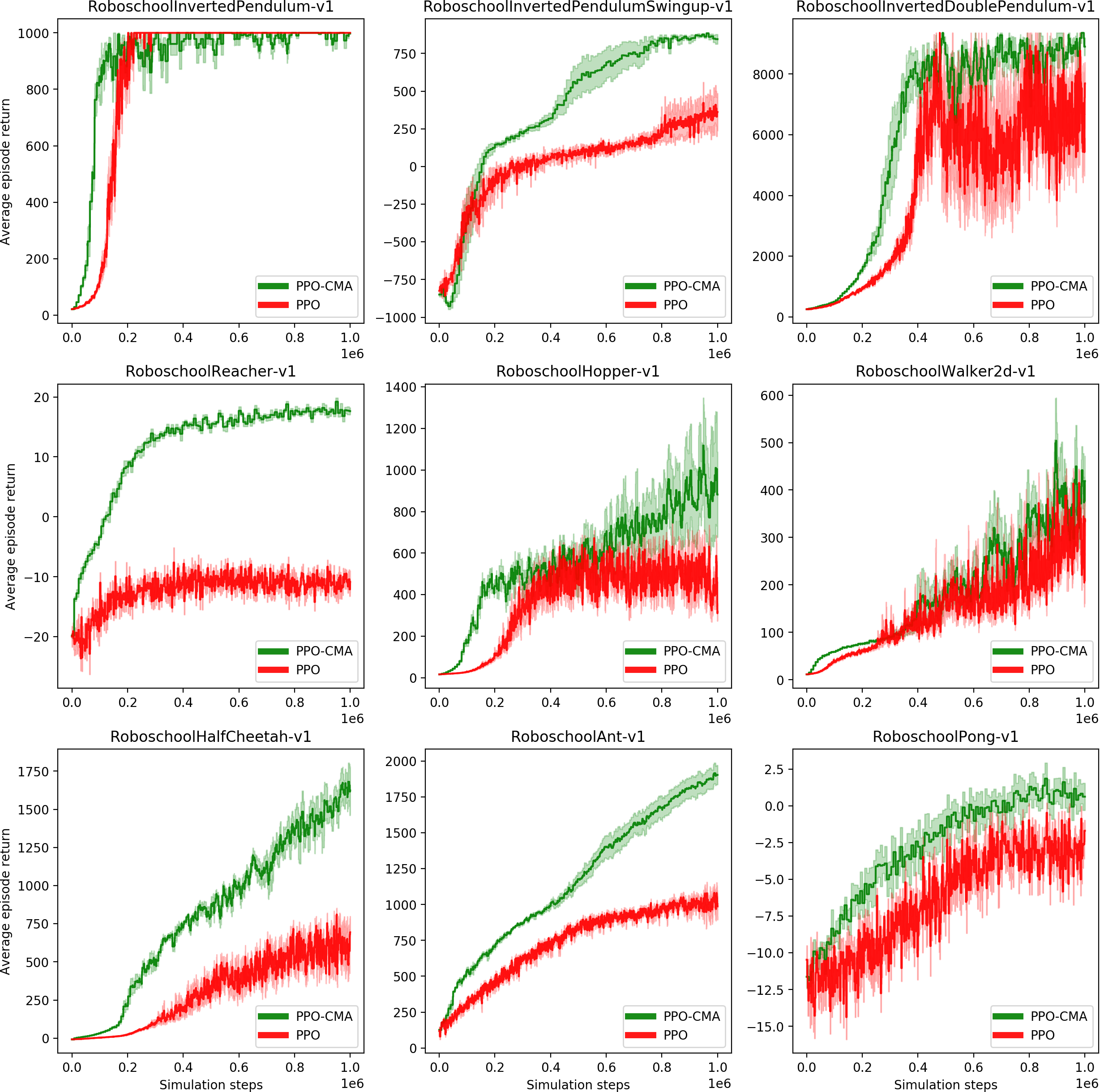}
\end{center}
\caption{Training curves from the 9 Roboschool environments used in the hyperparameter search. The plots use the best hyperparameter combinations in Figure \ref{fig:PPOvsPPO-CMA}.}\label{fig:ConvergencePlot}
\end{figure*}

Figure \ref{fig:ConvergencePlot} shows the training curves (mean and standard deviation of 5 runs) of all the 9 Roboschool environments used in the hyperparameter search. 

Overall, PPO-CMA performs clearly better in all environments except the inverted pendulum, where initial progress is slower. 

%% file: 00-Main.bbl
\begin{thebibliography}{10}

\bibitem{bergamin2019drecon}
Kevin Bergamin, Simon Clavet, Daniel Holden, and James~Richard Forbes,
\newblock ``Drecon: data-driven responsive control of physics-based
  characters,''
\newblock {\em ACM Transactions on Graphics (TOG)}, vol. 38, no. 6, pp. 1--11,
  2019.

\bibitem{peng2018deepmimic}
Xue~Bin Peng, Pieter Abbeel, Sergey Levine, and Michiel van~de Panne,
\newblock ``Deepmimic: Example-guided deep reinforcement learning of
  physics-based character skills,''
\newblock {\em arXiv preprint arXiv:1804.02717}, 2018.

\bibitem{schulman2017proximal}
John Schulman, Filip Wolski, Prafulla Dhariwal, Alec Radford, and Oleg Klimov,
\newblock ``Proximal policy optimization algorithms,''
\newblock {\em arXiv preprint arXiv:1707.06347}, 2017.

\bibitem{van2007reinforcement}
Hado van Hasselt and Marco~A Wiering,
\newblock ``Reinforcement learning in continuous action spaces,''
\newblock in {\em Approximate Dynamic Programming and Reinforcement Learning,
  2007. ADPRL 2007. IEEE International Symposium on}. IEEE, 2007, pp. 272--279.

\bibitem{babadi2018intelligent}
Amin Babadi, Kourosh Naderi, and Perttu H{\"a}m{\"a}l{\"a}inen,
\newblock ``Intelligent middle-level game control,''
\newblock in {\em 2018 IEEE Conference on Computational Intelligence and Games
  (CIG)}. IEEE, 2018, pp. 1--8.

\bibitem{levine2013guided}
Sergey Levine and Vladlen Koltun,
\newblock ``Guided policy search,''
\newblock in {\em International Conference on Machine Learning}, 2013, pp.
  1--9.

\bibitem{rajamaki2018regularizing}
Joose Rajam{\"a}ki and Perttu H{\"a}m{\"a}l{\"a}inen,
\newblock ``Regularizing sampled differential dynamic programming,''
\newblock in {\em 2018 Annual American Control Conference (ACC)}. IEEE, 2018,
  pp. 2182--2189.

\bibitem{abdolmaleki2018maximum}
Abbas Abdolmaleki, Jost~Tobias Springenberg, Yuval Tassa, Remi Munos, Nicolas
  Heess, and Martin Riedmiller,
\newblock ``Maximum a posteriori policy optimisation,''
\newblock {\em arXiv preprint arXiv:1806.06920}, 2018.

\bibitem{abdolmaleki2018relative}
Abbas Abdolmaleki, Jost~Tobias Springenberg, Jonas Degrave, Steven Bohez, Yuval
  Tassa, Dan Belov, Nicolas Heess, and Martin Riedmiller,
\newblock ``Relative entropy regularized policy iteration,''
\newblock {\em arXiv preprint arXiv:1812.02256}, 2018.

\bibitem{chua2018deep}
Kurtland Chua, Roberto Calandra, Rowan McAllister, and Sergey Levine,
\newblock ``Deep reinforcement learning in a handful of trials using
  probabilistic dynamics models,''
\newblock {\em arXiv preprint arXiv:1805.12114}, 2018.

\bibitem{schulman2015high}
John Schulman, Philipp Moritz, Sergey Levine, Michael Jordan, and Pieter
  Abbeel,
\newblock ``High-dimensional continuous control using generalized advantage
  estimation,''
\newblock {\em arXiv preprint arXiv:1506.02438}, 2015.

\bibitem{AWRPeng19}
Xue~Bin Peng, Aviral Kumar, Grace Zhang, and Sergey Levine,
\newblock ``Advantage-weighted regression: Simple and scalable off-policy
  reinforcement learning,''
\newblock {\em CoRR}, vol. abs/1910.00177, 2019.

\bibitem{schulman2015trust}
John Schulman, Sergey Levine, Pieter Abbeel, Michael Jordan, and Philipp
  Moritz,
\newblock ``Trust region policy optimization,''
\newblock in {\em International Conference on Machine Learning}, 2015, pp.
  1889--1897.

\bibitem{hansen2006cma}
Nikolaus Hansen,
\newblock ``The cma evolution strategy: a comparing review,''
\newblock in {\em Towards a new evolutionary computation}, pp. 75--102.
  Springer, 2006.

\bibitem{loshchilov2017limited}
Ilya Loshchilov, Tobias Glasmachers, and Hans-Georg Beyer,
\newblock ``Limited-memory matrix adaptation for large scale black-box
  optimization,''
\newblock {\em arXiv preprint arXiv:1705.06693}, 2017.

\bibitem{hansen2016cma}
Nikolaus Hansen,
\newblock ``The cma evolution strategy: A tutorial,''
\newblock {\em arXiv preprint arXiv:1604.00772}, 2016.

\bibitem{de2005tutorial}
Pieter-Tjerk De~Boer, Dirk~P Kroese, Shie Mannor, and Reuven~Y Rubinstein,
\newblock ``A tutorial on the cross-entropy method,''
\newblock {\em Annals of operations research}, vol. 134, no. 1, pp. 19--67,
  2005.

\bibitem{larranaga2001estimation}
Pedro Larra{\~n}aga and Jose~A Lozano,
\newblock {\em Estimation of distribution algorithms: A new tool for
  evolutionary computation}, vol.~2,
\newblock Springer Science \& Business Media, 2001.

\bibitem{hansen2010benchmarking}
Nikolaus Hansen and Raymond Ros,
\newblock ``Benchmarking a weighted negative covariance matrix update on the
  bbob-2010 noiseless testbed,''
\newblock in {\em Proceedings of the 12th annual conference companion on
  Genetic and evolutionary computation}. ACM, 2010, pp. 1673--1680.

\bibitem{OpenAI_roboschool}
OpenAI,
\newblock ``Roboschool,'' \url{https://blog.openai.com/roboschool/}, 2017.

\bibitem{brockman2016openai}
Greg Brockman, Vicki Cheung, Ludwig Pettersson, Jonas Schneider, John Schulman,
  Jie Tang, and Wojciech Zaremba,
\newblock ``Openai gym,''
\newblock {\em arXiv preprint arXiv:1606.01540}, 2016.

\bibitem{engstrom2019implementation}
Logan Engstrom, Andrew Ilyas, Shibani Santurkar, Dimitris Tsipras, Firdaus
  Janoos, Larry Rudolph, and Aleksander Madry,
\newblock ``Implementation matters in deep rl: A case study on ppo and trpo,''
\newblock in {\em International Conference on Learning Representations}, 2019.

\end{thebibliography}
